\documentclass[letterpaper]{article} 
\usepackage{aaai2026}  
\usepackage{times}  
\usepackage{helvet}  
\usepackage{courier}  
\usepackage[hyphens]{url}  
\usepackage{graphicx} 
\usepackage{natbib}  
\usepackage{caption} 
\usepackage{newfloat}
\usepackage{listings}
\DeclareCaptionStyle{ruled}{labelfont=normalfont,labelsep=colon,strut=off} 
\nocopyright

\usepackage{amssymb}            
\usepackage{mathtools}          
\usepackage{mathrsfs}           
\usepackage{graphicx}           
\usepackage{subcaption}         
\usepackage{url}                

\usepackage{algorithm}
\usepackage[noend] {algorithmic}
\usepackage{xspace}
\usepackage{tikz}
\definecolor{darkgreen}{RGB}{0,130,0}
\definecolor{lightpurple}{RGB}{147, 112, 219}
\newcount\Comments  
\long\def\shahaf#1{{\ifnum\Comments=1\color{purple}\bf [Shahaf: #1]\fi}}
\long\def\Gal#1{{\ifnum\Comments=1\color{magenta}\bf [Gal: #1]\fi}}

\newcommand{\astar}{\mbox{$A^*$}\xspace}

\newcommand{\start} {\mbox{$\mathit{start}$}\xspace} 
\newcommand{\goal} {\mbox{$\mathit{goal}$}\xspace}

\usepackage{acronym}

\acrodef{LHBL}{Limited-Horizon Bellman-based Learning}
\newcommand{\algname}{\ac{LHBL}\xspace}
\newcommand{\algnameonlysearch}{\ac{LHBL}$_S$\xspace}

\title{Beyond Single-Step Updates: Reinforcement Learning of Heuristics with Limited-Horizon Search}

\author{
    Gal Hadar\textsuperscript{\rm 1}
    Forest Agostinelli\textsuperscript{\rm 2}
    Shahaf S. Shperberg\textsuperscript{\rm 1}
}
\affiliations{
    \textsuperscript{\rm 1}Faculty of Computer and Information Science, Ben-Gurion University of the Negev\\
    \textsuperscript{\rm 2}Department of Computer Science and Engineering, University of South Carolina

}

\usepackage{bibentry}

\begin{document}

\maketitle  

\begin{abstract}
Many sequential decision-making problems can be formulated as shortest-path problems, where the objective is to reach a goal state from a given starting state. Heuristic search is a standard approach for solving such problems, relying on a heuristic function to estimate the cost to the goal from any given state. Recent approaches leverage reinforcement learning to learn heuristics by applying deep approximate value iteration. These methods typically rely on single-step Bellman updates, where the heuristic of a state is updated based on its best neighbor and the corresponding edge cost. This work proposes a generalized approach that enhances both state sampling and heuristic updates by performing limited-horizon searches and updating each state's heuristic based on the shortest path to the search frontier, incorporating both edge costs and the heuristic values of frontier states.
\end{abstract}

\section{Introduction}

Search algorithms have been fundamental to artificial intelligence (AI) since its inception, playing a central role in problem-solving and decision-making \citep{hart1968formal, bonet2001planning}. Among them, heuristic search, guided by a heuristic function to estimate the cost from a given state to a goal, has been particularly influential. It enables efficient navigation of vast solution spaces in domains such as route planning, hardware verification, theorem proving, robotics, and computational biology \citep{Edelkamp12}, many of which are often posed as reinforcement learning problems.

The effectiveness of heuristic search hinges on the accuracy of the heuristic function. While domain-specific heuristics can be highly effective, they often require expert knowledge. To address this limitation, domain-independent methods such as pattern databases \citep{culberson1998pattern} have been developed. However, these approaches involve trade-offs between computational cost and heuristic accuracy, especially as problem complexity increases \citep{agostinelli2019solving, muppasani2023solving}.

Recent advancements leverage deep neural networks (DNNs) \citep{schmidhuber2015deep} and reinforcement learning (RL) \citep{sutton2018reinforcement} to learn heuristics directly from data. Although these methods lack theoretical guarantees, they offer scalable, domain-specific heuristics that have demonstrated success in solving problems such as the Rubik’s Cube \citep{agostinelli2019solving}, chemical synthesis \citep{chen2020retro}, quantum algorithm compilation \citep{zhang2020topological, bao2024twisty}, and robotics \citep{tianmodel, eysenbach2019search}, often producing optimal or near-optimal solutions.

A common approach to learning heuristics through RL is approximate dynamic programming \citep{bellman1957dynamic, bertsekas1996neuro}. This involves sampling states from the environment and applying a single-step Bellman update, where a state's heuristic estimate is refined based on its best successor. DeepCubeA \citep{agostinelli2019solving} exemplifies this strategy, training a DNN via approximate value iteration to predict cost-to-go estimates, which are then used with a batched version of weighted A* search \citep{pohl1970heuristic} for efficient problem-solving.

In this paper, we demonstrate and tackle limitations in single-step Bellman-based heuristic updates. First, sampling random states in isolation neglects the inherent structure of search processes. In practice, states are not explored randomly but are instead closely related, forming local regions within the search space. Single-step updates fail to capture these dependencies when training a heuristic function. Additionally, single-step Bellman updates typically rely on the transition cost of a single edge---the edge to the best successor---while primarily depending on potentially inaccurate heuristic estimates of successors.

To address these challenges, we introduce Limited-Horizon Bellman-based Learning (LHBL), a method that performs limited-horizon searches from sampled states, leveraging the broader search context to generate training labels. Each state's heuristic is updated based on the best descendant on the frontier, combining the full path cost to the descendant and its heuristic for more informative labels than single-step updates. Additionally, since training states come from search, not random sampling, they better reflect those seen during actual search. This results in improved training efficiency and faster, more reliable search across multiple domains, as demonstrated in our empirical evaluation.


\section{Background and Related Work}
Pathfinding is a fundamental concept in AI with diverse applications \citep{hart1968formal, bonet2001planning}. A pathfinding problem instance, $I = (G, c, \start, \goal)$, where \( G = (V, E) \) is a graph with states (vertices) \( V \) and transitions (edges) \( E \subseteq V \times V \). The cost function \( c: E \rightarrow \mathbb{R}^{+} \) assigns a non-negative cost to each edge. The instance specifies a start state (\start) and either a goal state (\goal) or a goal predicate \( P: V \rightarrow \{0,1\} \), which indicates whether a state satisfies the goal condition. The objective is to find a path in \( G \) that connects \start to \goal with minimal cumulative cost, determined by the sum of the costs of its edges, though in some cases, finding a high-quality path quickly is prioritized over optimality, especially for large-scale problems or time-sensitive scenarios.

Heuristic search extends pathfinding by incorporating a heuristic function \( h: V \to \mathbb{R}^{+} \), which estimates the cost of the shortest path from a state \( s \) to the nearest goal state, commonly referred to as the \emph{cost-to-go}. The problem instance is then defined as $I = (G, c, \start, \goal, h).$
Heuristics can be domain-specific, like Manhattan distance in grid navigation, or derived automatically using techniques like pattern databases (PDBs) \citep{culberson1998pattern}, state-space transformations \citep{MostowP89}, and delete relaxations \citep{BonetG01}.

\subsection{Batch Weighted A*}

\astar \citep{hart1968formal} is a widely used and foundational search algorithm that maintains a priority queue, OPEN, of nodes discovered during the search, with each node containing state $n.s$, $g(n)$, the accumulated path cost from the start state to \( n \), and cost $f(n) = g(n) + h(n.s)$, where \( h(n.s) \) is the heuristic estimate of the remaining cost to a goal from state $n.s$. \astar iteratively removes and expands the node with the lowest cost ($f$-value), until selecting a node associated with a goal state for expansion. 

When A* uses a learned heuristic function (e.g., a deep neural network), heuristic computation can become a bottleneck. To mitigate this, GPUs can be leveraged to expand the \( B \) lowest-cost nodes in parallel, computing their heuristic values simultaneously. Still, A* can still be time- and memory-intensive. Weighted A* search \cite{pohl1970heuristic} mitigates this by adjusting the cost function:  
\begin{equation}  \label{eq:priority}
f(n) = \lambda g(n) + h(n.s),
\end{equation}
where \( \lambda \in [0,1] \) balances path cost and heuristic guidance. Lower 
$\lambda$ values bias toward heuristic-driven exploration, trading optimality for efficiency. Notably, the extreme case of $\lambda=0$ yields Greedy Best-first Search (GBFS), which abandons optimality to prioritize speed.

Combining parallel expansion and weighting yields \textbf{batch-weighted A* search (BWAS)}, a generalization of A* where standard A* is recovered by setting \( B = 1 \) and \( \lambda = 1 \). A* guarantees optimal solutions given an admissible heuristic, that is, a heuristic that never overestimates the cost of the shortest path: $h(s) \leq h^*(s)$ for all $s$, where $h^*(s)$ is the cost of a shortest path to a closest goal state from $s$  \citep{dechter1985generalized}. Similarly, weighted A* and BWAS guarantee bounded suboptimality under the same condition \cite{agostinelli2021obtaining, li2022optimal}. Although neural network-based heuristics are not necessarily admissible, research continues on developing admissible heuristics via deep learning \citep{ernandes2004likely, agostinelli2021obtaining, li2022optimal}. 

\subsection{Learning Heuristic Functions}
\label{sec:learning}
Research on learning heuristic functions has explored methods to enhance heuristic-based algorithms for over three decades. One approach is imitation learning, where cost-to-go values, often derived from domain knowledge or solvers, are used to train heuristics via supervised learning. \citet{samadi2008compressing} reduced memory usage by training neural networks to approximate Pattern Database (PDB) heuristics. Other studies have designed architectures for heuristic learning in planning problems~\citep{abs-2112-01918,TakahashiSTW19,ShenTT20,FerberH020,ToyerTTX20}, proposing alternative loss functions to improve search efficiency~\citep{GarrettKL16,BhardwajCS17,GroshevGTSA18,abs-2310-19463}. These approaches are constrained by their reliance on pre-existing solvers or expert solutions, which are often unavailable for many real-world problems.

Another approach uses reinforcement learning (RL) to learn the heuristic function from the costs of paths found using heuristic search. \citet{Bramanti-GregorD93} iteratively refined heuristics with \astar and trained new heuristics via linear regression. \citet{Fink07} learned a weighted sum of admissible heuristics, while \citet{ArfaeeZH11} used neural networks and random walks to generate easier instances when no problems were solved. \citet{OrseauL21} extended this by learning both a policy and a heuristic. A major limitation of these methods is their inability to learn from expanded nodes that do not contribute to a solution, leading to poor sample efficiency.

A recent direction in heuristic function learning involves leveraging large language models (LLMs) for heuristic generation. For instance, \citet{ling2025complex} proposes an automated heuristic discovery method that uses LLMs to derive heuristic functions for complex planning tasks without requiring explicit domain knowledge. Similarly, \citet{correa2025classical} demonstrates that heuristics generated by LLMs can rival established classical planning heuristics; however, their approach still relies on encoding domain-specific details within prompts, thereby maintaining a dependence on domain knowledge. Moreover, the practical effectiveness and scalability of these approaches in diverse planning problems remain uncertain, necessitating further empirical validation.

A notable alternative method separates the learning from the search process, using approximate dynamic programming \citep{bellman1957dynamic, bertsekas1996neuro} via single-step Bellman (SSB) updates \citep{bellman1957dynamic}. This technique iteratively refines heuristic estimates by randomly sampling a state $s \in V$ (either directly or by taking random moves from $\goal$) and updating the estimate based on the minimum transition cost plus the estimated cost-to-go for its neighbors, following the second pathmax rule~\citep{Mero84}:
\begin{equation} \label{eq:bellmanup} 
h_{SSB}(s) = \min_{s'\in V \text{ s.t. } (s,s')\in E}{c(s,s') + h(s')} \end{equation}
Iterative Bellman updates are effective for approximating the true cost-to-go~\citep{bertsekas1996neuro}. \citet{ThayerDR11} applied them during search, using \( h_B(s) \) as the ground truth to improve the heuristic. 

DeepCubeA~\citep{agostinelli2019solving} applies iterative Bellman updates to train a deep neural network (DNN) heuristic, denoted \( h_\theta \), where \( \theta \) are the DNN parameters, to minimize Bellman error across domains. For each state \( s \), the updated heuristic is computed as:
\begin{equation} \label{eq:bellmanup_fa} 
\small
h_{SSB}(s) =
\begin{cases}
0, & \text{if } s = \goal, \\
\min\limits_{s' \in V \text{ s.t. } (s,s') \in E}{\left( c(s,s') + h_{\theta^-}(s') \right)}, & \text{otherwise}.
\end{cases}
\end{equation}
Here, \( \theta^- \) denotes the parameters of a \emph{target network}~\citep{MnihKSGAWR13}---a slower-updating copy of the main network used to compute stable target values and reduce oscillations during training.
The DNN is trained by minimizing the mean squared error (MSE) between the updated and predicted heuristic estimates:
\begin{equation} \label{eq:trainloss}
L(\theta) = \frac{1}{N}\sum_{i=1}^N \left(h_{SSB}(s_i) - h_{\theta}(s_i)\right)^2.
\end{equation}
To solve problems, the learned heuristic is used within BWAS.
DeepCubeA is the first machine-learning approach to reliably solve various puzzles, including the Rubik's cube, 35-puzzle, and Lights Out without human guidance. Its success has inspired methods for challenges in quantum computing \citep{zhang2020topological,bao2024twisty}, cryptography \citep{jin20203d}, and chemical synthesis \citep{chen2020retro}.

\section{Limited-Horizon Heuristic Update}







Single-step Bellman-based learning (SSBL) has empowered search algorithms to solve previously intractable problems---an especially remarkable achievement considering it requires no domain knowledge or example solutions. However, SSBL has two key limitations.

First, in SSBL, training examples are generated by sampling from the environment, either by directly modifying state variables or by taking random backward moves from the goal. The motivation behind this approach is to learn heuristic estimates from states distributed uniformly across the entire state space, allowing generalization across different problem instances with varying start states.
However, while a problem instance can start from any state in the state space, the distribution of states encountered during search is far from uniform. Nodes are expanded in order based on the priority function (Eq.~\ref{eq:priority}), which ranks nodes according to the cost of the path from the start state and their heuristic estimate. Although the distance from the start state ensures that states are sampled across the space in expectation, the heuristic function introduces a non-uniform bias that varies between problem instances and states.

Specifically, the search process favors nodes with lower heuristic values, leading to a significant overrepresentation of states with lower heuristic estimates across all problem instances. This effect is even more pronounced for smaller values of $\lambda$, which place greater emphasis on heuristic values over path costs---peaking at the extreme case of Greedy Best-First Search (GBFS), which considers only heuristic estimates. This difference in state distribution during training and deployment can lead to heuristics that suffer from large depression regions, where the heuristic function significantly under-
estimates the true cost-to-go.

The second limitation of SSBL is its limited use of known, accurate information in each update step. As formalized in Equation~\ref{eq:bellmanup_fa}, each heuristic update for a state \( s \) estimates the distance to the goal by considering the edge cost \( c(s, s') \) plus the heuristic estimate from \( s' \) to the goal, for each neighboring state \( s' \). The final heuristic update takes the minimum of these values.  
Notably, only the edge costs are exact, while the heuristic of \( s' \) remains an approximation. As a result, each heuristic update, which aims to estimate a complete path cost from \( s \), incorporates only a single known edge cost, with the rest of the path cost being estimated.  

In the tabular setting, where each state's heuristic is stored independently and updated exactly, this process is guaranteed to converge to the optimal cost-to-go, albeit potentially requiring many iterations. In contrast, when using function approximators (e.g., neural networks), relying heavily on uncertain estimates can degrade performance. Leveraging more accurate information in each update—beyond a single edge—could lead to more reliable and efficient learning.

To address these limitations, we introduce \algname, a method for state sampling and updating based on limited-horizon search. \algname consists of two phases: \emph{search} and \emph{update}. In the search phase, an initial state \(s\) is sampled from the environment, similar to SSBL. However, rather than immediately performing a Bellman update and sampling a new state, \algname conducts a search from \(s\) for a fixed number of expansions \(N\). This search can be performed using any algorithm (e.g., \astar, GBFS), providing flexibility in exploration. The heuristic guiding the search is derived from the target network. The search phase of \algname ensures that the distribution of states encountered during training aligns with those encountered when deploying the heuristic to solve problems.

\begin{figure}[tb]
    \centering
    \includegraphics[trim=25 11 17 1, clip, width=\linewidth]{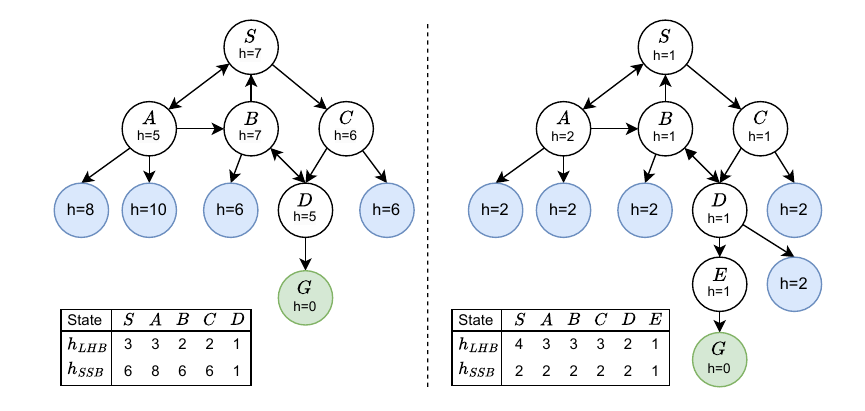}
    \caption{Graph examples for comparing \( h_{LHB} \) and \( h_{SSB} \).}
    \label{fig:graph_backup}
\end{figure}

Beyond improving the distribution of sampled states, incorporating search during training enables more accurate heuristic estimation. Given a partially expanded search graph \( G \), and a node \( n \in G \), any path from \( n.s \) to the goal must pass through one of its descendant leaf nodes. Thus, the optimal path must also pass through one of them.

Rather than relying solely on immediate successors---i.e., a single edge cost and estimated heuristic values---we leverage the full subtree rooted at \( n \), using complete path costs and descendant heuristics for more informed updates.
This idea parallels multi-step lookahead in RL--e.g., \textit{n}-step SARSA~\citep{de2018multi}, which generalizes one-step TD learning by bootstrapping value estimates after \( n \) steps along the trajectory taken by the agent to accelerate convergence and reduce variance. These RL methods can accelerate convergence and reduce variance. We extend this approach by performing updates over entire descendant trees.

Formally, we define the set of descendants of a node \( n \) in the search graph \( G \) as:
\[
D(n) = \{ \ell \in G \mid \ell \text{ is reachable from } n \text{ in } G, \ell \neq n \}.
\]  
The subset of descendant nodes that are leaves is given by  
\[
L(n) = \{ \ell \in D(n) \mid \ell \text{ has no children in } G \}.
\]  
For each leaf \( \ell \in L(n) \), let \( P(n, \ell) \) denote the shortest path from \( n \) to \( \ell \), represented as a sequence of states  
\[
P(n, \ell) = (s_0, s_1, \dots, s_k), \quad \text{where } s_0 = n.s \text{ and } s_k = \ell.s.
\]  
The accumulated path cost to \( \ell \) is then given by  
\[
C(n, \ell) = \sum_{i=0}^{k-1} c(s_i, s_{i+1}).
\]  

Instead of using a single-step Bellman update, the heuristic estimate is updated as  
\begin{equation} \label{eq:bellmanup_LH} 
h_{LHB}(s) =
\begin{cases}
0, & \text{if } s = \goal, \\
\min\limits_{\ell \in L(n)} \left( C(n, \ell)+ h_{\theta^-}(\ell.s) \right), & \text{otherwise}.
\end{cases}
\end{equation}

where $LHB$ stands for Limited-Horizon Bellman.
This approach ensures that the heuristic considers a more informed estimate by incorporating the cost along the best reachable path within the limited-horizon search, rather than relying mostly on the heuristic estimations. 

Figure~\ref{fig:graph_backup} illustrates two examples highlighting the advantage of \algname (Eq.~\ref{eq:bellmanup_LH}) over SSBL (Eq.~\ref{eq:bellmanup_fa}). In these examples, white circles represent states expanded during the limited-horizon search, the search frontier is shown in blue, and the goal state, \( G \), is depicted in green. Each node contains its prior heuristic estimate, and all edge costs are set to 1.  
The leftmost figure demonstrates a scenario where \( h_{SSB} \) overestimates the cost-to-go, while \( h_{LHB} \) correctly learns an accurate estimate in a single update. Consider state \( S \) in this example. SSBL evaluates the heuristic estimates of its immediate successors \( A \), \( B \), and \( C \), selecting the minimum estimate plus the edge cost. Since \( A \) yields the lowest value, the update results in \( h_{SSB}(S) = 6 \). In contrast, LHB considers all leaf nodes in the search frontier (blue and green nodes). The node yielding the minimum value under Eq.~\ref{eq:bellmanup_LH} is \( G \), leading to \( h_{LHB}(S) = 3 \), which accounts for the path cost from \( S \) to \( G \) and the heuristic value of \( G \) (which is zero).  
The right graph in Figure~\ref{fig:graph_backup} presents a different scenario, where \( h_{SSB} \) underestimates the cost-to-go, whereas \( h_{LHB} \) produces a more accurate estimation.



\subsection{Computing Limited-Horizon Bellman Updates}
\begin{figure}[t]
    \centering
    \includegraphics[trim=19 8 17 1, clip, width=\linewidth]{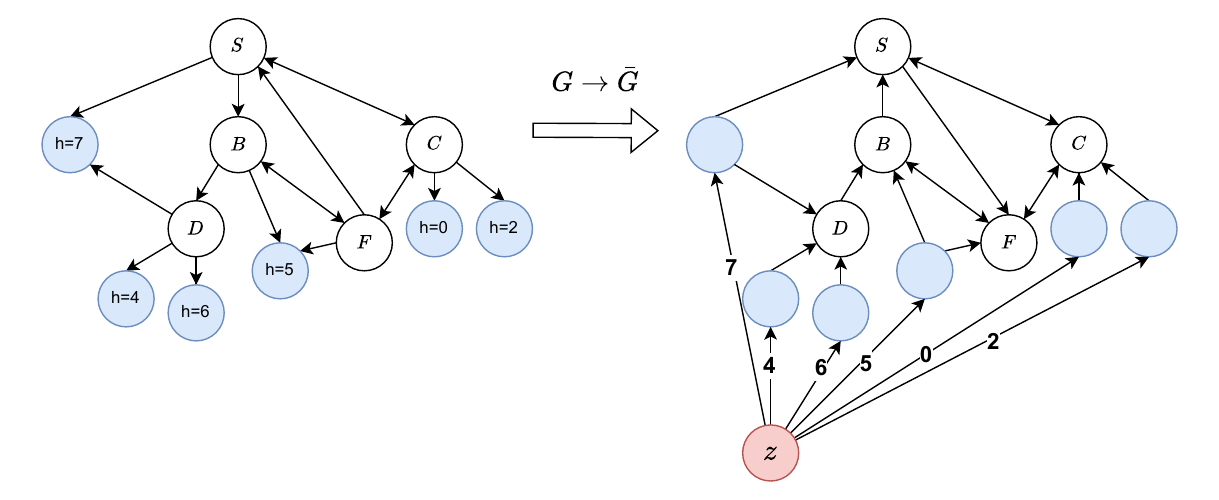}
    \caption{Illustration of graph transformation for computing $h_{LHB}$. The blue nodes are the frontier of the search graph; the red state $z$ is the auxiliary state.}
    \label{fig:graph_transform}
\end{figure}

\begin{algorithm}[t]
\caption{Limited-Horizon Bellman-based Learning}
\label{alg:LHG}
\begin{algorithmic}[1]
\REQUIRE State $s$, expansion budget steps $N$, heuristic function $h$
\ENSURE Updated heuristic values $h_{LHB}$
\STATE \textbf{Step 1: Construct Search Graph}  
\STATE Run a search algorithm (e.g., A*, GBFS) with $s$ as the start node for $N$ steps to obtain search graph $G(V, E)$.
\STATE \textbf{Step 2: Add auxiliary Node}  
\STATE Introduce an auxiliary node $z$ and define: $\bar{V} \leftarrow V \cup \{z\}$, $\bar{E} \leftarrow E$.
\FORALL{$\ell \in G$ such that $\ell$ is a leaf node}
    \STATE Add directed edge $\bar{E} \leftarrow \bar{E} \cup \{(\ell,z)\}$ with cost $c(\ell,z) = h(\ell)$.
\ENDFOR
\STATE \textbf{Step 3: Reverse Graph Edges}  
\STATE $\bar{E} \leftarrow \{(v,u) | (u,v) \in \bar{E}\}$
\STATE \textbf{Step 4: Compute Heuristic Values}  
\STATE Run Dijkstra’s algorithm from $z$ on graph $\bar{G} = (\bar{V},\bar{E})$.
\FORALL{$v \in V$}
    \STATE Assign heuristic value $h_{LHG}(v)$ as the shortest path distance in $\bar{G}$ from $z$ to $v$.
\ENDFOR
\end{algorithmic}
\end{algorithm}

Computing \( h_{LHB} \) for a state \( n.s \) involves finding the frontier node \( \ell \) that minimizes the path cost from \( s \) to \( \ell \) plus \( \ell \)'s heuristic estimate.

A natural approach to computing \( h_{LHB} \) might be a recursive application of single-step Bellman updates, starting from the leaves and propagating updates upward to the root while using the updated estimates of successors at each step. However, this method does not account for cycles that may exist in the partially expanded graph \( G \). To address this issue, we propose reducing the computation of \( h_{LHB} \) for each state to a single-source shortest path (SSSP) problem.  

To achieve this, we construct a modified graph \( \bar{G} \) by introducing an auxiliary node \( z \), as illustrated in Figure~\ref{fig:graph_transform}. This auxiliary node is connected to all frontier nodes via incoming edges, where each edge cost is determined by the heuristic value of the state represented by the corresponding frontier node. In this modified graph \( \bar{G} \), the shortest path from any node \( n \) to \( z \) corresponds to the cost of reaching the best frontier node in \( G \) from \( n \).  

To efficiently compute the shortest path from every node to \( z \), we reverse the direction of all edges in \( \bar{G} \) and solve the SSSP problem from \( z \) using Dijkstra’s algorithm \citep{dijkstra1959note}. This transformation ensures an efficient and cycle-aware computation of \( h_{LHB} \). 
The entire process of generating training examples in \algname, based on a given (randomly sampled) state $s$ is summarized in Algorithm \ref{alg:LHG}.

Notably, our limited-horizon mechanism is conceptually similar to real-time search methods like LSS-LRTA* \cite{koenig2009comparing}, as both use local search to inform heuristic updates rather than single-step backups. The contexts, however, are distinct: LSS-LRTA* is an online, tabular method for real-time action, whereas our Limited-Horizon Bellman-based Learning (LHBL) is an offline procedure for generating labels to train a deep neural network. Furthermore, LSS-LRTA* assumes consistent h-values, resulting well-behaved, monotonic online update. We make no such assumption, as our DNN heuristic $h_{\theta-}$ is not guaranteed to be consistent. Therefore, we require the more general SSSP formulation, which uses a reversed graph and auxiliary node to ensure a cycle-aware computation of the $h_{LHB}$ target label.

\begin{figure*}[tbp]
    \centering
    \includegraphics[width=\linewidth]{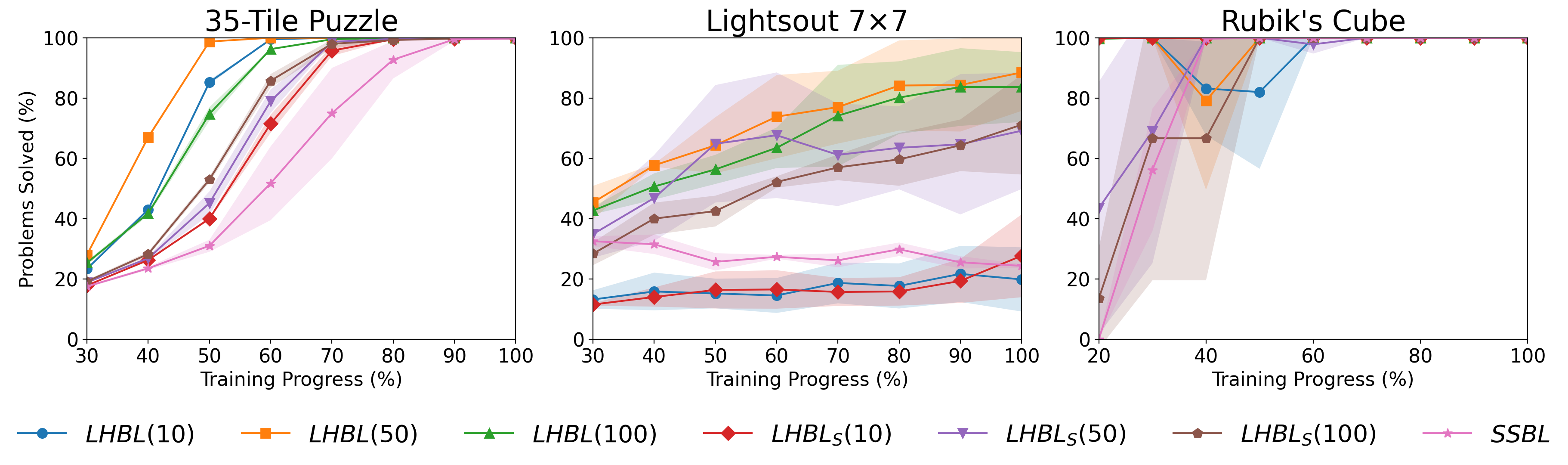}
    \caption{Problems solved throughout the training: STP, LightsOut, and Rubik's Cube.}
    \label{fig:training}
\end{figure*}
\section{Empirical Evaluation}



This section presents an empirical evaluation comparing \algname\ to SSBL. Our goal is to analyze the two key contributions of \algname: (1) leveraging search to obtain a more representative distribution of training examples and (2) using LHB (Eq.~\ref{eq:bellmanup_LH}) to compute improved heuristic estimates.



\subsection{Evaluation Settings}






To implement our approach, we extended the DeepCubeA framework \citep{deepcubeagithub} with \algname.\footnote{The code will be made public upon acceptance.}
We evaluated the approach on three puzzle domains: Rubik's Cube, 35 Sliding Tile Puzzle (STP), and $7\times 7$ Lights Out, which are described in Appendix A.

We ran seven different algorithms on each domain. The first algorithm, SSBL, generates each training example by sampling a state from the environment and applying the single-step Bellman update (Eq.~\ref{eq:bellmanup_fa}).
The second algorithm, \algnameonlysearch, uses limited-horizon search for generating training examples while retaining the standard single-step Bellman update (Eq.~\ref{eq:bellmanup_fa}) for assigning labels. This variant allows us to isolate and empirically examine the impact of modifying the training sample distribution independently of the update method. We evaluated three search horizons, 10, 50, and 100, resulting in \algnameonlysearch$\!(10)$, \algnameonlysearch$\!(50)$, and \algnameonlysearch$\!(100)$.
Finally, we ran the full \algname algorithm, combining limited-horizon search for training sample generation and LHB (Eq.~\ref{eq:bellmanup_LH}) for computing training labels. We evaluated search horizons of 10, 50, and 100, resulting in \algname$\!(10)$, \algname$\!(50)$, and \algname$\!(100)$.



To reduce experimental noise and ensure a focused and fair comparison of the update methods, all experiments were performed using the same neural network architecture and hyperparameters described in the DeepCubeA paper \citep{agostinelli2019solving}. Consistency across all methods was further maintained by employing identical sequences of randomly sampled states during both training and testing phases. During the training phase, Rubik’s Cube and STP-35 domains were trained on approximately 10 billion generated samples, whereas the LightsOut domain was trained on approximately 1 billion samples. During the training phase, all algorithms were trained on the same identical sequences of randomly sampled states to allow a fair convergence analysis. The Rubik's Cube and STP-35 domains generated approximately 10 billion samples, whereas the LightsOut domain generated approximately 1 billion samples. For the testing phase, we used the first 200 instances from the test set of problem instances used in the DeepCubeA paper \citep{agostinelli2019solving}. Each instance in the test set was generated with a varying number of random steps from the goal state to assess different levels of instance difficulty. To account for stochastic variability, each experimental setup was independently repeated three times with different random seeds, and the results are reported as the mean and standard deviation calculated over these runs.

Training was conducted on an Nvidia RTX 4090 GPU with 60GB of RAM, while experiments were performed on an RTX 3090 GPU with 50GB of RAM. Training times varied by domain: for the LightsOut domain, each algorithm and seed required approximately 3--4 days; for the STP, training took 6--7 days; and for the Rubik’s Cube domain, training ranged from 10 to 12 days.

\begin{figure*}[tb]
    \centering
    \resizebox{0.97\textwidth}{!}{
    \includegraphics[width=\linewidth]{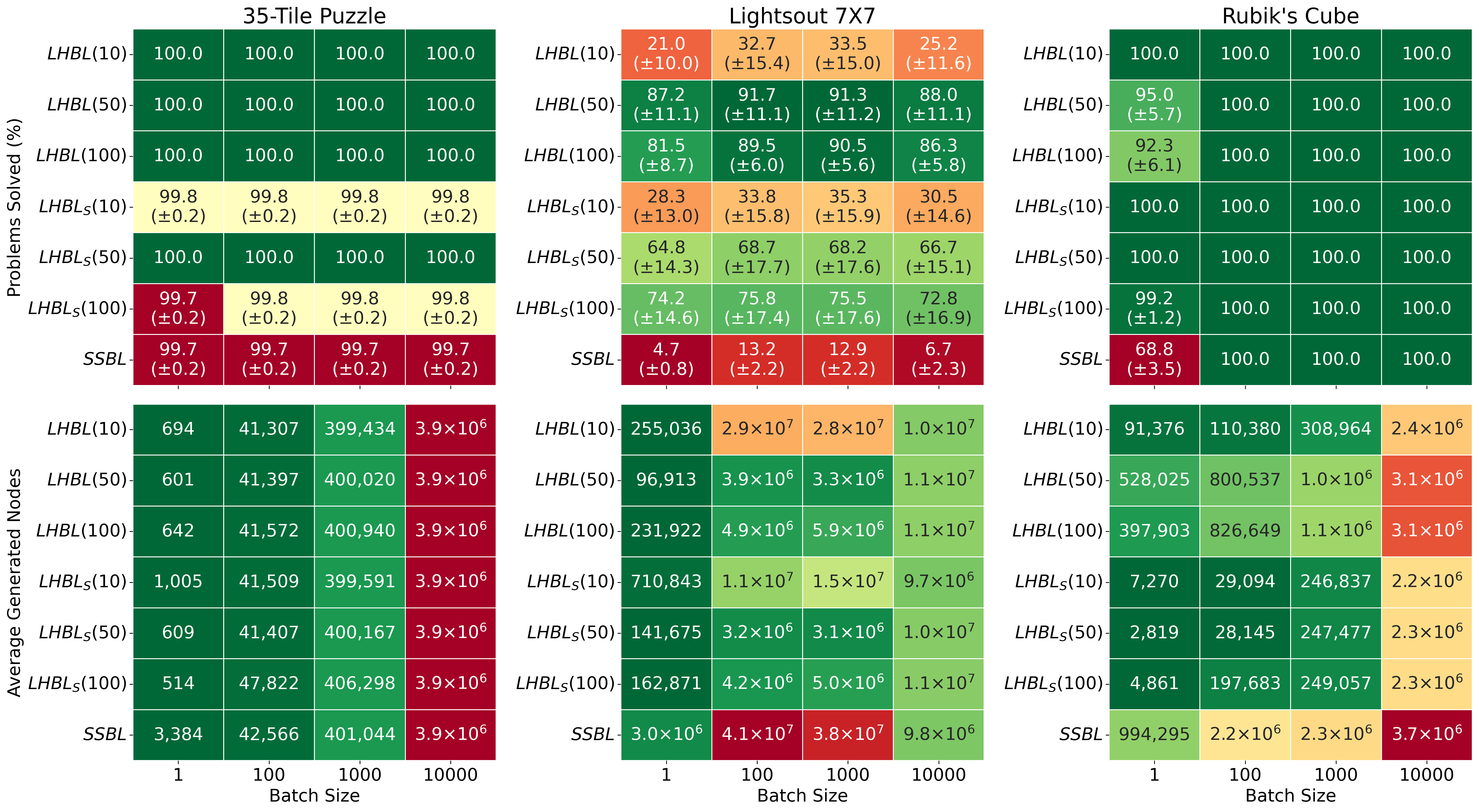}
    }
    \caption{Results on fully trained heuristic: STP, LightsOut, and Rubik's Cube.}
    \label{fig:full_percentage}
\end{figure*}
\subsection{Results on Training Progress}

In this experiment, we stored checkpoints of the learned heuristics at different stages of training and evaluated them by using the heuristic as part of BWAS in an attempt to solve all testing instances. A time limit of 10 minutes was set for each problem instance. To ensure consistency with DeepCubeA, we adopted its original test configuration, using domain-specific batch sizes: 20,000 for the 35-STP, 10,000 for the Rubik's Cube, and 1,000 for LightsOut. 

Figure~\ref{fig:training} reports the training progress of each method. 
In each plot, the y-axis depicts the percentage of problems solved, whereas the x-axis shows the percentage of training progress. Solid lines represent the mean, and the shaded areas represent the standard deviation between the three seeds.

The results on STP indicate that SSBL exhibits the slowest training among all approaches, gradually improving but never reaching a 100\% success rate, even by the end of training. In contrast, all other approaches successfully solve all problem instances after completing at most 80\% of the training. Furthermore, the \algname variant consistently outperforms the \algnameonlysearch variants, with each \algname variant dominating its corresponding \algnameonlysearch variant throughout the entire training process. Additionally, in this domain, \algname benefits slightly more from a search horizon of 10 compared to 100, whereas the opposite trend is observed for \algnameonlysearch.  

In the LightsOut domain, \algname$\!(100)$ and \algname$\!(50)$ solved the most problem instances, showing steady improvement throughout the search. In contrast, SSBL, \algname$\!(10)$, and \algnameonlysearch$\!(10)$ performed significantly worse, with little improvement over the course of training. However, as we will see next, the fully trained heuristics obtained by \algname$\!(10)$ and \algnameonlysearch$\!(10)$ significantly outperform those of SSBL. Notably, across all configurations, \algname variants outperform their corresponding \algnameonlysearch variants.

The results on the Rubik's Cube domain show that the early stages of training are quite noisy, even when averaging over three random seeds, making it difficult to draw clear conclusions. Nevertheless, all algorithms converged to solving all problem instances by 70\% of the training process. Notably, \algname$\!(100)$ consistently solved all instances much earlier---by just 20\% of the training.

Overall, most \algname and \algnameonlysearch variants require fewer states 
to be sampled from the environment to solve a high percentage of problem instances, indicating faster convergence through improved sample efficiency.


\begin{figure}[tb]
    \centering
    \hspace{1.2mm}
    \begin{minipage}{0.45\textwidth}
        \centering
        \begin{tikzpicture}
        \node[anchor=south west,inner sep=0] (image) at (-0.18,0) {\includegraphics[width=0.5\linewidth]{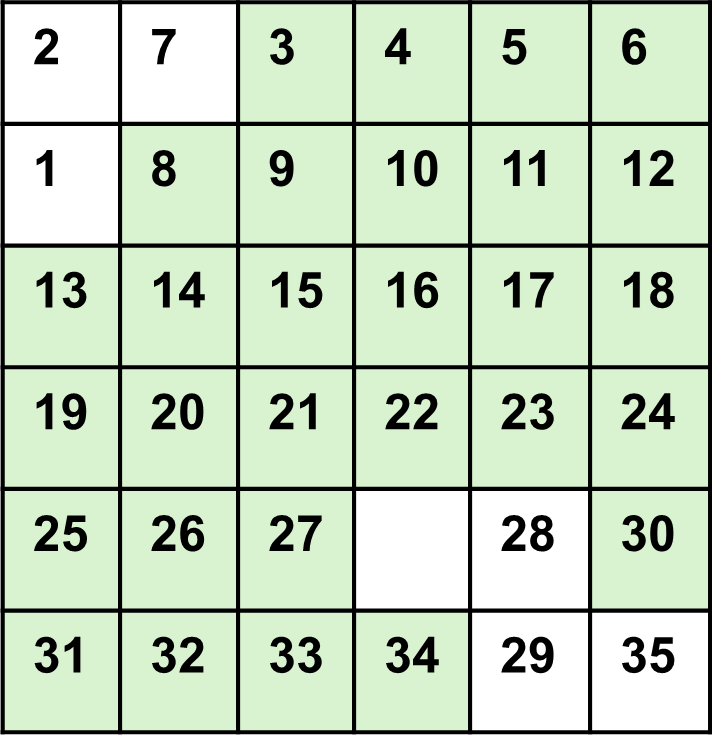}};
        \node[black, fill=white, opacity=0.7, text opacity=1] at (-1.8,3.5) {\footnotesize \algname$\!(\!10) = 15.7$}; 
        \node[black, fill=white, opacity=0.7, text opacity=1] at (-1.8,3.0) {\footnotesize \algname$\!(\!50) = 15.3$}; 
        \node[black, fill=white, opacity=0.7, text opacity=1] at (-1.8,2.5) {\footnotesize \algname$\!(\!100) = 14.6$}; 
        \node[black, fill=white, opacity=0.7, text opacity=1] at (-1.8,2.0) {\footnotesize \algnameonlysearch$\!(\!10) = 16.3$}; 
        \node[black, fill=white, opacity=0.7, text opacity=1] at (-1.8,1.5)
        {\footnotesize \algnameonlysearch$\!(\!50) = 16.2$}; 
        \node[black, fill=white, opacity=0.7, text opacity=1] at (-1.8,1.0){\footnotesize \algnameonlysearch$\!(\!100)\!=\!14.6$}; 
        \node[black, fill=white, opacity=0.7, text opacity=1] at (-1.8,0.5) {\footnotesize SSBL $= 3.1$}; 
        \end{tikzpicture}
        \caption{Example of STP representative state inside a depression region. Most tiles are in the correct place, but the state is still many steps away from being solved.}
        \label{fig:stp:example}
    \end{minipage}
\end{figure}

\begin{figure*}[tbp]
    \centering
    \begin{subfigure}{0.33\textwidth}
        \includegraphics[width=\linewidth]{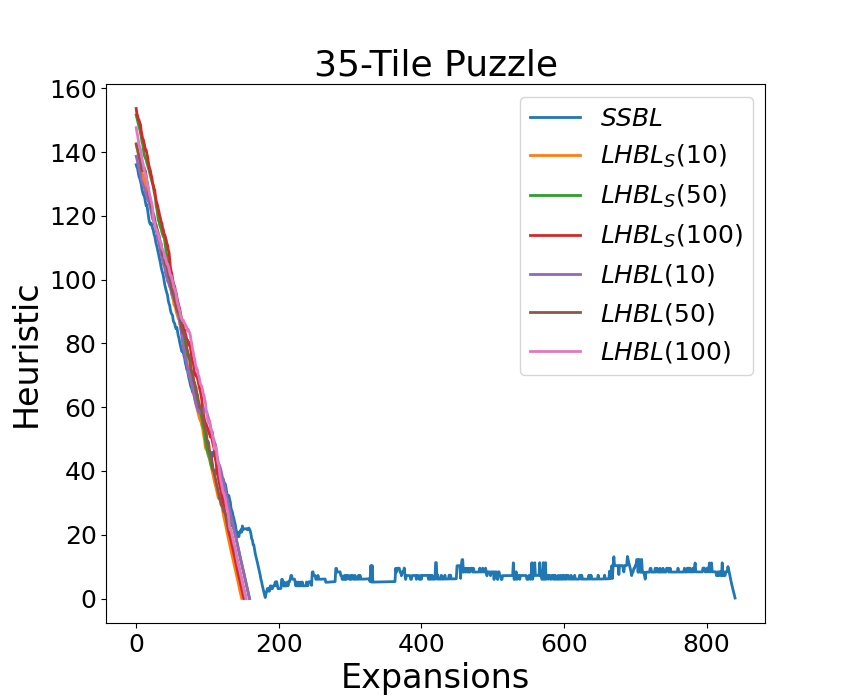}
    \end{subfigure}
    \hspace{-1.5mm}
    \begin{subfigure}{0.33\textwidth}
        \includegraphics[width=\linewidth]{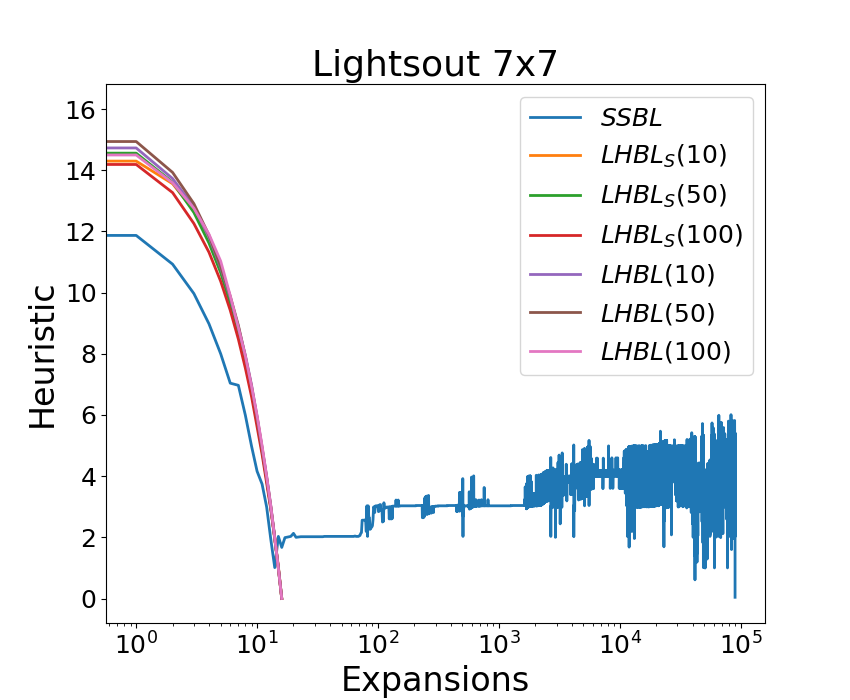}
    \end{subfigure}
    \hspace{-1.5mm}
    \begin{subfigure}{0.33\textwidth}
        \includegraphics[width=\linewidth]{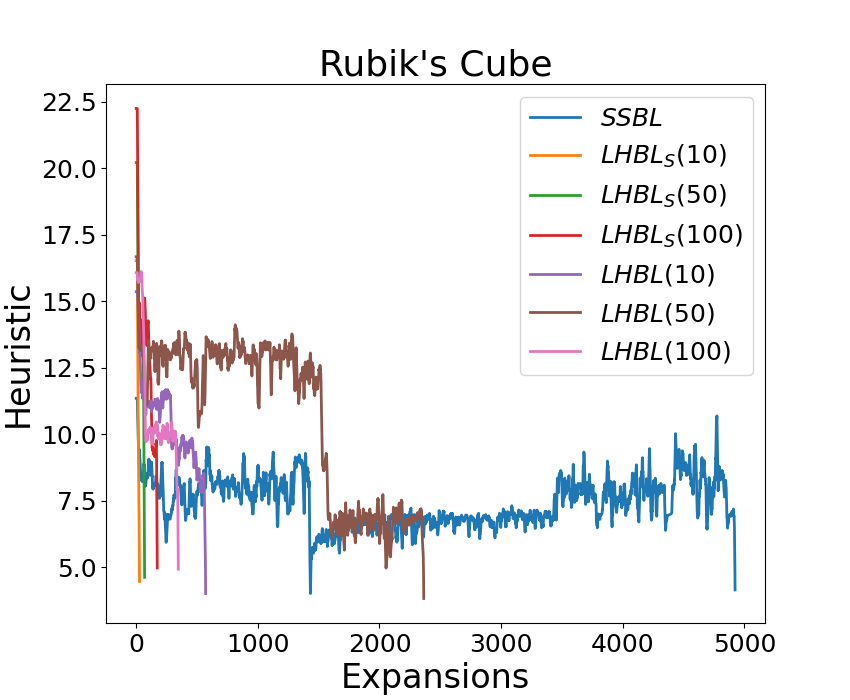}
    \end{subfigure}
    \caption{Depression region examples, many expansions with relatively low heuristic values indicate the algorithm is in a depression region. \algname and \algnameonlysearch mitigates these depression regions.}
    \label{fig:depression}
\end{figure*}

\subsection{Results on the Fully-trained Heuristics}
In this experiment, we focus exclusively on the fully trained heuristics obtained by each algorithm while broadening the scope of the evaluation. For each heuristic, we ran the full test set of problem instances using BWAS with a fixed weight of $\lambda=0.6$, and batch sizes of 1, 100, 1,000, and 10,000. We maintain a 10-minute time limit per problem instance, consistent with our previous experiment. 
The results are presented as heatmaps, illustrating the performance of each method across different batch sizes, metrics, and domains. Each cell in the heatmap contains the average value for a given configuration, with the standard deviation (computed across seeds) shown in parentheses. Figure~\ref{fig:full_percentage} presents the results for the percentage of problems solved and the average number of node generations. We also compared runtime; however, since all network architectures are identical, runtime is effectively captured by the number of node generations. Therefore, we include the runtime and results only in the appendix.

In STP, all algorithms solved all instances except one, which SSBL failed to solve and \algnameonlysearch$\!(100)$ and \algnameonlysearch$\!(10)$ struggled with in some seeds.
Examining node generation, we observe several trends. First, node generations increase with batch size. This aligns with expectations, as larger batch sizes cause more nodes to be expanded before new ones are added to the search frontier, slowing search progression. Second, the performance gap between algorithms narrows as the batch size increases. This is expected, as heuristic quality has a greater impact when the batch size is small; with a sufficiently large batch size, the search effectively degenerates into breadth-first search. At a batch size of 1, where heuristics play the most significant role, SSBL exhibits substantially higher node generation than search-based training approaches.

In Rubik's Cube, all algorithms were able to solve all instances for batch sizes greater than one. However, for a batch size of one, \algnameonlysearch$(\!100)$, \algname$(\!100)$, \algname$(\!50)$, and SSBL were the only algorithms that failed to solve all instances, with SSBL performing the worst, solving only 68.8\% of the problems—significantly fewer than all other approaches.
This disadvantage is further reflected in SSBL’s average node generation and runtime, which are two orders of magnitude higher than those of the \algnameonlysearch variants when using a batch size of 1. Moreover, this performance gap persists across all batch sizes. Notably, in this domain, the \algnameonlysearch variants consistently outperformed the \algname variants across all configurations. 

In the LightsOut domain, SSBL consistently underperforms relative to the other approaches, solving only 4.7\% to 13.2\% of the instances across all batch sizes. Across all configurations, the \algname variants outperformed the \algnameonlysearch variants, with the sole exception of \algnameonlysearch\!(10).


\subsection{Examples of Depression Regions}

As previously shown, SSBL expands more nodes and solves fewer problems—particularly at batch size 1. This is likely due to heuristic depression regions \citep{aine2016multi}, where the heuristic significantly underestimates the true cost-to-go. These regions can lead to inefficient search and, if large enough, exhaust available memory. Learned heuristics are especially prone to this issue, as such regions often occupy small, hard-to-sample parts of the state space.

Figure~\ref{fig:depression} shows search progress on a representative instance from each domain, plotting heuristic values (y-axis) against the number of states expanded (x-axis) for all algorithms. Across all domains, the results clearly demonstrate that SSBL encounters large depression regions during search, whereas the limited-horizon search significantly mitigates these regions, improving search efficiency.

To better understand states in depression regions, we sampled several such states. These states appear close to the goal in state-variable values but require many more steps to reach it. A representative example from the STP domain is shown in Figure~\ref{fig:stp:example}, where most tiles are correctly placed, and misplaced tiles have a low Manhattan distance to their target positions. However, due to movement constraints---where tiles can only be shifted into the blank space---many moves are still needed. LHBL’s heuristic more accurately reflects the difficulty of these states, while SSBL misestimates them as close to the goal, leading to inefficient search.

\section{Summary and Conclusion}

In this work, we introduced \algname, a reinforcement learning-based method for learning heuristics that improves upon single-step Bellman updates by leveraging limited-horizon search. Our approach enhances the state sampling distribution during training and refines heuristic estimates by considering entire paths to the search frontier rather than just immediate successors. By formulating the heuristic update process as a single-source shortest path problem, \algname efficiently integrates longer-term dependencies into heuristic learning while mitigating the impact of depression regions.  

Our empirical evaluation demonstrated that \algname generally outperforms standard single-step Bellman-based learning (SSBL). In particular, \algname improves sample efficiency, leads to faster training convergence, and results in more effective heuristics when used within search algorithms. Increasing the search horizon during training improves heuristic quality by incorporating deeper lookahead, but overly large horizons can introduce approximation errors or overfit to specific deep paths.


\section*{Acknowledgements}
The work of Shahaf Shperberg was supported by the Israel Science Foundation (ISF)
grant \#909/23,  by Israel's Ministry of Innovation, Science and Technology (MOST) grant \#1001706842, in collaboration with Israel National Road Safety Authority and Netivei Israel, awarded to Shahaf Shperberg, by BSF grant \#2024614 awarded to Shahaf Shperberg, as part of a joint NSF-BSF grant with Forest Agostinelli. This material is based upon work supported by the National Science Foundation under Award No. 2426622.

\bibliography{main}

\newpage

\onecolumn

\appendix
\section*{\LARGE
Supplementary Materials for the ``Beyond Single-Step Updates: Reinforcement Learning of Heuristics with
Limited-Horizon Search" Paper}

\vspace{1cm}

\section{Appendix A: Domain description}
\vspace{0.3cm}
 
The Rubik's Cube is a 3D combination puzzle consisting of a $N \times N \times N$ cube, where each of the six faces is covered by nine stickers of a single color. The cube can be manipulated by rotating any of its six faces, which moves rows or columns of smaller cubelets. The goal is to return the cube to its solved state, where each face shows a uniform color, starting from a scrambled configuration.
The Sliding-tile puzzle (STP) is a combinatorial puzzle consisting of a square grid with $N$ numbered tiles and an empty space, which allows tiles to be moved to the empty space. The goal is to arrange the tiles in ascending numerical order from 1 to $N$. 
The Lights Out game is a classic puzzle consisting of a $N \times N$ grid of lights, where each light can be in either an ON or OFF state. At the start, a random subset of lights is switched on, and clicking on a light toggles its state along with its adjacent lights. The goal is to turn all lights off using the fewest possible moves.  For our experiments, we used $N = 3$ for the Rubik's Cube ($4.3\cdot10^{19}$ unique states), $N = 35$ for the Sliding Tile Puzzle ($3.72 \cdot 10^{41}$ possible arrangements), and $N = 7$ for Lights Out ($10^{14}$ unique states), ensuring sufficiently complex problem instances for evaluating our approach. 

\section{Appendix B: Additional Results}
\vspace{0.3cm}

This section presents the runtime results of the experiment using fully trained heuristics. Overall, the runtime trends align with the node generation patterns reported earlier. Notably, there is a tradeoff between batch size and runtime. While processing large batches on the GPU is significantly more efficient than computing heuristics for individual states, larger batch sizes also lead to increased node generations. Overall, a batch size of 100 achieves the best average runtime across all algorithms and domains.

\begin{figure*}[h]
    \centering
    \includegraphics[width=\linewidth]{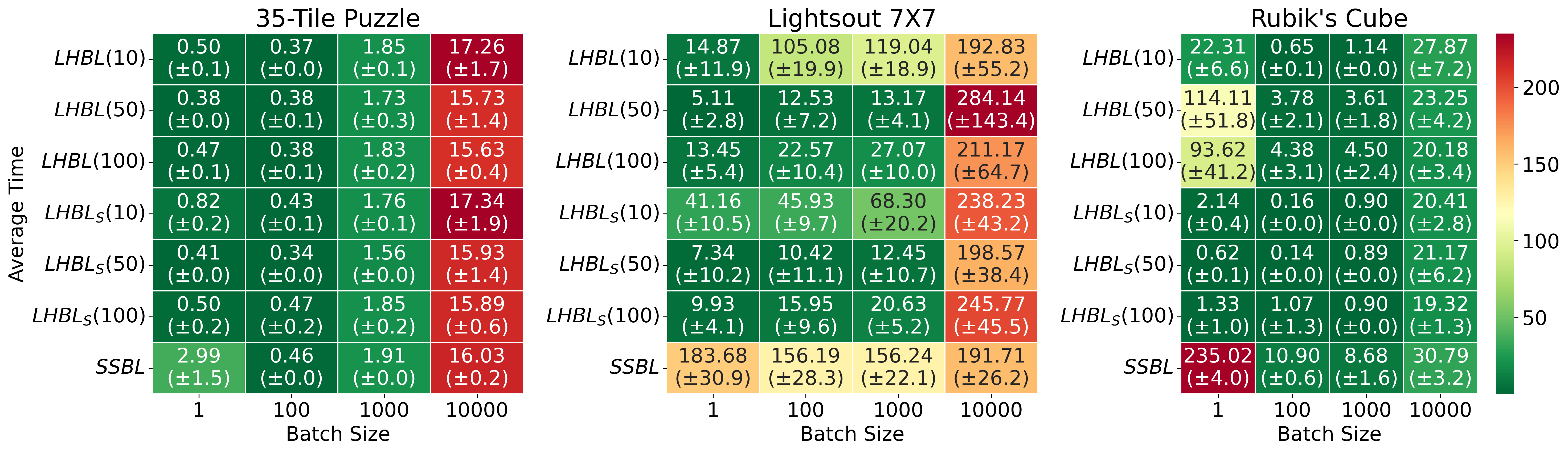}
    \caption{Runtime on fully trained heuristic: STP, LightsOut, and Rubik's Cube.}
    \label{fig:full_percentage_additional}
\end{figure*}

\end{document}